\documentclass[letterpaper, 10 pt, conference]{ieeeconf}  

\IEEEoverridecommandlockouts                              

\usepackage{amsmath} 
\usepackage{amssymb}  
\usepackage{amsfonts}
\usepackage[table]{xcolor}
\definecolor{Mycolor1}{HTML}{DAE8FC}
\definecolor{Mycolor2}{HTML}{FFE6CC}
\definecolor{Mycolor3}{HTML}{D5E8D4}
\usepackage{soul}
\usepackage{graphicx}
\usepackage{makecell}
\usepackage{float}
\usepackage{graphics}
\usepackage{etoolbox}
\usepackage{booktabs}
\usepackage{hyperref}
\usepackage{url}
\usepackage{svg}
\usepackage{subcaption}
\usepackage{algorithm}
\usepackage{algpseudocode}

\title{\LARGE \bf
CoT-TL: Low-Resource Temporal Knowledge Representation of Planning Instructions Using Chain-of-Thought Reasoning

}

\author{Kumar Manas$^{\star,\dagger}$, Stefan Zwicklbauer$^{\dagger}$ and Adrian Paschke$^{\star,\ddagger}$
\thanks{$^{\star}$Freie Universit{\"a}t Berlin, Germany. $^{\dagger}$Continental Automotive Technologies GmbH, Germany. $^{\ddagger}$Fraunhofer FOKUS, Germany. 
\newline
This work is partially funded by the German Federal Ministry for Economic Affairs and Climate Action within the project KI Wissen.
 \newline
        {Email:\tt\small   kumar.manas [at] fu-berlin [dot] de}
        }
}

\usepackage{textcomp}
\usepackage[absolute,showboxes]{textpos}
\TPMargin{5pt}

\begin{document}

\maketitle
\thispagestyle{empty}
\pagestyle{empty}

\begin{abstract}
Autonomous agents often face the challenge of interpreting uncertain natural language instructions for planning tasks. Representing these instructions as Linear Temporal Logic (LTL) enables planners to synthesize actionable plans. We introduce CoT-TL, a data-efficient in-context learning framework for translating natural language specifications into LTL representations. CoT-TL addresses the limitations of large language models, which typically rely on extensive fine-tuning data, by extending chain-of-thought reasoning and semantic roles to align with the requirements of formal logic creation. This approach enhances the transparency and rationale behind LTL generation, fostering user trust. CoT-TL achieves state-of-the-art accuracy across three diverse datasets in low-data scenarios, outperforming existing methods without fine-tuning or intermediate translations. To improve reliability and minimize hallucinations, we incorporate model checking to validate the syntax of the generated LTL output. We further demonstrate CoT-TL's effectiveness through ablation studies and evaluations on unseen LTL structures and formulas in a new dataset. Finally, we validate CoT-TL's practicality by integrating it into a QuadCopter for multi-step drone planning based on natural language instructions. Project details: \href{https://github.com/kumarmanas/TAMP\_COT\_TL}{https://github.com/kumarmanas/TAMP\_COT\_TL}
\end{abstract}

\section{Introduction}\label{intro}
\begin{figure}[ht]
    \centering
    \includegraphics[width=7.4cm, height=4.0cm]{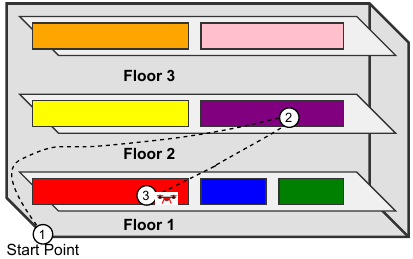}
    \caption{Illustration of LTL formula \texttt{$F\bigl(purple\_room \wedge F(red\_room)\bigr)$} from natural language specification \textit{Move to purple room, before moving to the red room} being used for drone planning. The dotted black line shows the drone’s path.}
    \label{fig:teaser}
    \vspace{-18pt}
\end{figure}

Autonomous agents, such as robots and drones, need to follow natural language instructions and rules for planning. Natural language can be vague, inconsistent, or assume prior knowledge. Temporal logic, including its variant linear temporal logic (LTL), is a formal language used to clearly and precisely represent instructions and rules for agents, ensuring that their actions comply with these directives~\cite{LTL_satisfy}. LTL adds temporal operators to propositional logic. For example, a natural language rule for a drone could be ``go to the purple room, then go to red room”. The LTL specification of this rule is \texttt{$F\bigl(purple\_room \wedge F(red\_room)\bigr)$}, which means that the drone should visit the purple room and then eventually to the red room. Agents need to understand this planning task to reach the goal (Fig.\ref{fig:teaser}), and LTL has been shown to be a good representation choice~\cite{dronetestfine,lang2ltl}. However, the manual creation of LTL specifications is time-consuming for most users unfamiliar with the logic syntax, semantics, and domain knowledge. This problem can limit the adaptation of LTL in planning. A semantic parser for automated formalization of natural language specification to LTL can assist the process. 

In this paper, we treat automated formalization as an LTL translation task, where natural language instructions is translated into LTL using a large language model (LLM) without relying on a large fine-tuning dataset. We use LLMs for their state-of-the-art translation performance~\cite{gpt3,gpt4}. However, LLMs need to be fine-tuned using annotated datasets to create such semantic parsers. Dataset creation is costly as it involves collection effort and logic experts. Moreover, the end tasks for autonomous agents vary in nature and only sometimes allow us to have a reasonable and adequate dataset for good performance and generalization. 

Prompting, as introduced in~\cite{gpt3}, requires very few data samples (few-shot learning) or sometimes no data samples (zero-shot learning). However, they perform poorly on tasks requiring reasoning~\cite{wei2023chainofthought}, essential for a more accurate representation of LTL. We present a framework, \textbf{CoT-TL} to address this. Our framework extends the chain-of-thought (CoT) prompting to translate natural language planning specifications into LTL formulas from known atomic propositions (APs). Additionally, we improve translation accuracy by using semantic role labeling (SRL) to align LLMs more effectively with domain-specific tasks, improving the accuracy of LTL formula generation. CoT-TL generates interpretable output for the end user, which enables us to visualize the LLM’s reasoning process behind LTL generation and offers opportunities for further debugging and improvement. Such a framework facilitates communication between the user and autonomous agent and ensures that the agent’s planner adheres to the instructions safely and is trusted. 

Previous semantic parsers often require large datasets~\cite{copynet} or rely on synthetic data~\cite{dronedataset} in the absence of large data, limiting their adaptability to new environments~\cite{robotdataset}. Our training-free method outperforms existing approaches in accuracy, data efficiency, and robustness without large amounts of data, intermediate translation steps, or data preprocessing. Intermediate translation means converting specifications into simpler forms by paraphrasing or converting them into canonical forms. We leverage the world knowledge of LLMs and only need a few high-quality data samples as few-shot examples. Our method can effectively learn from a few detailed but relevant data samples (cf.~\ref{sub:prompts}). Additionally, CoT-TL ensures that the generated LTL formula is parsable (correct syntax) and satisfiable through model checking (cf.~\ref{constrain}). Our key contributions include:
\begin{itemize}
    \item Training free framework to translate natural language planning instructions into LTL using LLMs, without large fine-tuning datasets with competitive performance
    \item Systematic approach to extend CoT reasoning for synthesizable temporal specification creation
    \item Generation of interpretable and end-user-friendly temporal specification showing reasoning steps involved, enabling safety and trust
\end{itemize}

\section{Releated Work}
\label{related_work}
LTL representation is used to generate executable code for agent planning. Previous works have shown the feasibility of generating code directly from a natural language without using a formal representation~\cite{cop,progprompt}. These approaches struggle with multiple instances of the same object, ensuring a formal guarantee of code correctness, optimality, and adaptation to different control primitives. They also complicate the debugging of the failures. Other works use physical demonstrations to learn temporal specifications~\cite{chou_learning_2022,bayesdemo}, but such approaches are time consuming and costly. Therefore, we aim to use natural language to LTL to make such methods accessible to a wide range of non-expert end users. Another set of previous work~\cite{mcmahon06, artzi13weakly, chen11} based on executing language-guided robot tasks has focused on semantic parsing to ground language commands into abstract representations capable of informing robot actions. 

Before the advent of deep learning, semantic parsing and SMT solvers were used to create LTL specifications from natural language in robotics planning~\cite{semparseltl}. The limited performance of classical semantic parsers and template techniques has motivated using LLMs. The robotics domain also uses the transformer~\cite{attention} architecture for verification, planning, and reasoning tasks~\cite{llmasplanner, llmrobot}. 

A predefined template translates the text in natural language into temporal logic in \cite{nl2spec,tr2mtl} and~\cite{lang2ltl}. Work by~\cite{lstmtranslator,robotdataset} uses domain-specific models to convert natural language specifications into LTL. Prompting is explored in~\cite{lang2ltl} with some success. To address the data scarcity problem, an intermediate canonical form translation is introduced to help the semantic parsing problem~\cite{canoform}. The works most related to ours are~\cite{dronetestfine} and Lang2LTL~\cite{lang2ltl}, which use neural network-based models and LLMs to translate specifications into LTL formulas. However, these works use synthetic or real data with intermediate translations to generate the LTL formula. This makes them less applicable in low-data scenarios or requires additional data preprocessing. Lang2LTL uses intermediate referring expression representation of text specifications and uses lifted translation alongside prompting, but the main focus of their work is fine-tuning. In contrast, our work does not rely on intermediate translations or fine-tuning of the model. Work by~\cite{liftedgen,copynet} enhances semantic parser's generalization to a new domain, which is close and complementary to our work, but they need intermediate translation and processing. Similarly, Cook2LTL~\cite{cook2ltl} used semantic parsers for instruction preprocessing and caching mechanisms to reduce LLM calls.
\section{Preliminaries}
\label{prelim}
\subsection{Linear Temporal Logic (LTL)}
\label{sub:LTL}
We informally introduce the operators used in LTL based on~\cite{ltl1} and~\cite{ltl2}. LTL specifications used in this work can be written with below LTL grammar :
\begin{equation}
\varphi ::= p\mid \neg p \mid \varphi_1 \wedge \varphi_2 \mid \varphi_1 \vee \varphi_2 \vspace{-0.5mm}
\end{equation}
Where $p \in P$ and $P$ is a set of possible boolean APs corresponding to the environmental state where autonomous agents may or may not navigate, $\varphi$ is the task specification, $\varphi_1$ and $\varphi_2$ are LTL formulas. The logical connectives \textit{negation} ($\neg$), \textit{and} ($\wedge$) and \textit{or} ($\vee$) are used to write formulas. We also have the following temporal operators:
\begin{equation}
    \varphi ::= \text{G}(\varphi) \mid \varphi_1 \text{U}\varphi_2 \mid \text{F}(\varphi) \vspace{-0.7mm}
\end{equation}
where $G$, $U$ and $F$ are temporal operators. The future globally operator $G$ specifies that $\varphi$ holds within a time interval for all future states. The until operator $U$ specifies that $\varphi_1$ holds until $\varphi_2$ becomes true. The future operator or eventual operator $F$, specifies that $\varphi$ holds within a time interval for some future state.
\subsection{Large Language Model and Chain-of-Thought}
\label{sub:LLM}
Employing the notation $p_\theta$ to represent a pre-trained LLM distribution with parameters $\theta$, and a language sequence $x=(x[1], \cdots, x[n])$ where each $x[i]$ is a token or words, so that $p_\theta(x) = \prod_{i=1}^{n} p_\theta(x[i] | x[1...i])$ generate the next token probability distribution. Here, we explain various strategies that leverage such LLMs for problem-solving tasks. 
\textbf{Input-output (IO) prompting} uses LLM to produce an output ($y$) for input task ($x$) by adding task instructions to the input, and output is generated as $y \sim p_\theta^{IO}(y|x)$. However, this method struggles with problems that require multiple reasoning steps. 

\begin{figure}[ht]
    \centering
    \includegraphics[width=8.2cm, height=4.3cm]{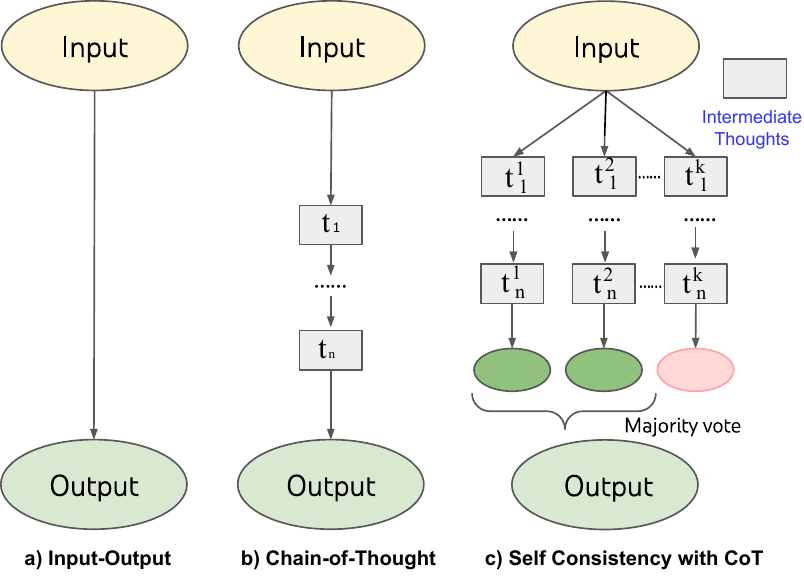}
    \caption{Illustration of different in-context learning prompts. }
    \label{fig:prompts}
    \vspace{-12pt}
\end{figure}

To overcome this, \textbf{Chain-of-thought (CoT) prompting}~\cite{wei2023chainofthought} was introduced. This is particularly useful when the mapping of $x$ and $y$ is non-trivial (e.g., $x$ is textual rule and $y$ is formal logic). It involves chaining a series of intermediate language sequences, called ``thoughts," $t_1, \cdots, t_n$ to bridge the gap between $x$ and $y$. Each thought is generated by the LLM in sequence $t_i \sim p_\theta^{CoT}(t_i \mid x, t_{1\cdots i-1})$ and represents a meaningful step toward problem solving $y \sim p_\theta^{CoT}(y | x, t_{1 \cdots n})$. It is similar to breaking down complex mathematical problems into smaller steps and solving each step before deriving the final answer. This reduces errors, and we can implicitly access the model reasoning process. CoT works by giving a few examples of how to solve a problem using a CoT and then asking the LLM to follow the same pattern for a new problem. CoT excels at capturing various intricate reasoning methods that surpass the abilities of IO prompting. However, CoT prompting struggle with compositional generalization task (understanding unseen combinations of seen primitives)~\cite{compo_gen}. In our context, if the model learned to generate $a \wedge b$ and $a \wedge c$ through prompting, it should ideally be capable of generalizing to $a \wedge b \wedge c$ in the presence of compositional generalization.
\textbf{Self-consistency with CoT (CoT-SC)}~\cite{Wang2022SelfConsistencyIC} prompting improves LLMs robustness. In CoT-SC, LLMs generate multiple independent chains of thoughts. CoT-SC involves extracting $k$ i.i.d. chains of thought: $[t^{(i)}_{1\cdots n}, y^{(i)}] \sim p_\theta^{CoT}(t_{1\cdots n}, y | x) \ (i=1 \cdots k)$, then returns the most frequent output (majority voting): $\arg \max_{y} \#\{i\mid y^{(i)}=y\}$. This i.i.d sampling of $k$ thoughts enhances the reliability and robustness of output translation. This work extends CoT-SC for the formal logic domain.
Fig.~\ref{fig:prompts} illustrates the above-mentioned three prompting techniques.

\subsection{Problem Definition}
\label{sub:problem}
We frame our problem as the task of translating natural language specifications $\mathcal{U}$ into LTL formula $\varphi$ for autonomous agents in the low data domain. Later, $\varphi$ is synthesized by an automaton for planning. LTL translation is the main focus of this work, and we leverage the intrinsic in-context learning capability of LLMs for this. We assume that our framework CoT-TL has access to the list of possible APs and that the grounding of APs for autonomous agents is known. We leverage CoT and semantic parsing, specifically SRL, which complement our prompting approach to generate accurate LTL formula generation without using the intermediate translation of natural language specification $\mathcal{U}$. In summary, we aim to generate mapping $\mathcal{U} \mapsto \varphi$ instead of $\mathcal{U} \mapsto C \mapsto \varphi$ as in~\cite{dronetestfine,lang2ltl} through LLMs, where $C$ can be an intermediate translation or simplified paraphrase of $\mathcal{U}$.
\section{CoT-TL: Architecture}
\label{methodology}
\begin{figure*}[t]
    \centering
    \includegraphics[width=\textwidth]{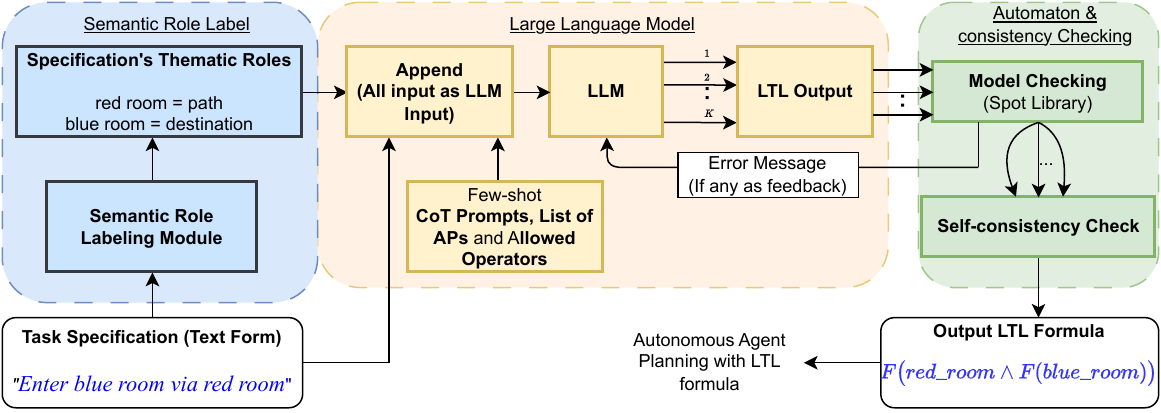}
    \caption{\textbf{CoT-TL:} LTL formula generation pipeline. Blue blocks represent the SRL module, Yellow indicates the LLM module, and green blocks represent the output constraining and automaton feasibility checking module.}
    \label{figs:arch}
    \vspace{-14pt} 
\end{figure*} 
\label{arch}
CoT-TL is a framework to generate an LTL formula $\varphi$ from natural language specification $\mathcal{U}$ without an additional intermediate translation module and fine-tuning the LLM. Our architecture is shown in Fig.~\ref{figs:arch}. It has 3 submodules: (a) SRL (in blue) to identify the thematic roles of the words in $\mathcal{U}$, (b) an LLM module (in yellow) that uses SRL information, CoT prompts, and automaton checking feedback, and (c) an model checking (in green) that can parse and verify the correctness of the LLM generated LTL formulas alongside self-consistency check. For simplicity, we explain each module using specification $\mathcal{U}$ as ``Enter blue room via red room" and respective formula $\varphi$ as \texttt{$F\bigl(red\_room \wedge F(blue\_room)\bigr)$}.

\subsection{Semantic Role Labeling (SRL)}
\label{sub:SRL}
SRL identifies the underlying semantic structure within text specification by analyzing the relationships between words and phrases. This is achieved by finding predicate (verbs), arguments (entity), and contextual aspects such as time and location from the specification. \textbf{SRL facilitates semantic parsing—the process of translating natural language into formal representations} like LTL—by clarifying the \textit{predicate-argument structure}. SRL adeptly manages the variability inherent in natural language, as the same expressions can be expressed in multiple ways, but in the end, they may have similar semantic meanings. In low-data contexts, this is particularly advantageous. Our approach is based on VerbNet~\cite{verbnet} and~\cite{manassrl} for semantic role identification, assigning thematic roles to phrases to use them in prompt, and aiding in detecting atomic propositions (APs) for LTL formula generation. For example, in ``Enter blue room via red room", CoT-TL needs to identify the parts of the sentence that correspond to the APs. In this case, ``blue room” and ``red room” are two such parts. SRL helps CoT-TL in this identification. We assign the thematic role of \textit{destination} to blue room and the thematic role of \textit{path} to via red room. This information guides the LLMs in identifying APs and the agent’s action sequence in the specification, and we use SRL information as a soft constraint for the LLMs by using them as part of the overall prompt fed to the LLM. Refer~\cite{verbnet} for more details.\\
\textbf{SRL module input}: Enter blue room via red room. \\
\textbf{SRL module output}: {Enter [arg 0] blue
room [destination] via red room [path].}

\subsection{Chain-of-Thought and Few-shot Prompts}
\label{sub:prompts}
We utilize few-shot prompting instead of fine-tuning to show CoT-TL usability in the limited data domain. In IO prompting~(\ref{sub:LLM}), the LLMs are provided with an instruction and output pair as: ``Convert the specification into LTL: Enter blue room via red room" and ``\texttt{$F\bigl(\texttt{red\_room} \wedge F(\texttt{blue\_room})\bigr)$}". We can have $N$ such pairs, known as $N$-shot prompting. Inspired by CoT~\cite{wei2023chainofthought}, CoT-TL provides an intermediate thought process and step-by-step breakdown of complex problems into subgoals. Due to subgoal creation, LLMs can focus on small individual steps and reduce distraction induced by multi-step specification. Such reasoning skills are needed for temporal logic, as it requires knowledge about formal logic as well as an understanding of natural language semantics, which can be vague and assume background knowledge about the domain. Our CoT-TL differentiates itself from~\cite{wei2023chainofthought} by introducing a CoT methodology tailored for formal logic generation. In contrast, the previous CoT work~\cite{wei2023chainofthought} focused primarily on mathematical and question-answering tasks, with limited consideration of formal logic. Formal logic tasks require a combination of natural language semantics and adherence to additional constraints imposed by the formal language grammar. Moreover, in LTL logic generation, the emphasis lies more on understanding temporal action sequences and disambiguating potentially vague or context-dependent instructions rather than solely on rigorous reasoning and theorem understanding. Since LLMs pre-training corpus does not explicitly include a high proportion of formal logic, unlike code generation, translation, or question-answering tasks~\cite{gpt3,mistral,Li2023StarCoderMT}, we need a new approach to CoT prompting, where we also leverage SRL with CoT. By assigning semantic roles to the words in the specification, SRL (\ref{sub:SRL}) assists in the translation task for CoT-TL.

\begin{figure}[ht!]
    \centering
    \includegraphics[width=\linewidth]{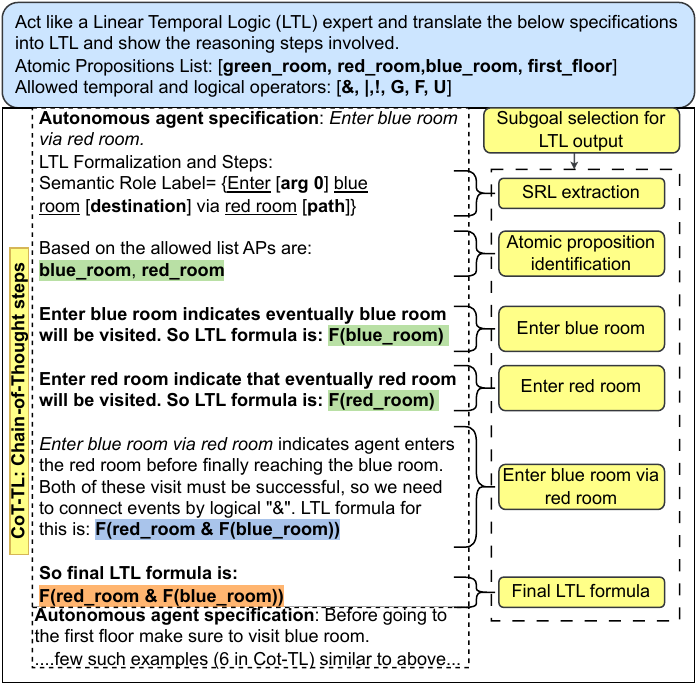}
    \caption{\textbf{CoT-TL prompting methodology:} CoT-TL prompt file includes a header (blue box) defining the main instruction, allowed APs, and operators. It then contains a series of CoT prompts designed for LTL tasks, each with a few-shot example pair showing the derivation of LTL from the specification. Our prompts design breaks down the specifications into subgoals. The model solves these subgoals sequentially, using earlier answers to reason about later ones.The final subgoal's answer triggers LTL generation. \textbf{During testing}, we append the header and 6 CoT prompts examples with test time specification, and the CoT-TL produces an interpretable LTL output similar to the prompt pattern. \textit{Apart from annotations in yellow boxes all other texts are part of prompt file}.} 
    \label{fig:myprompt} \vspace{-12pt}
\end{figure}

CoT-TL extends the classical CoT~\cite{wei2023chainofthought} further for formal specification. Fig.~\ref{fig:myprompt} shows our prompt creation strategy. This approach mitigates ambiguity and addresses the compositional generalization problem. This approach has shown to be particularly helpful in determining the correct nesting of temporal and logical operators in our evaluation. As shown in Fig.~\ref{fig:myprompt}, SRL information is used along with other components of the prompts (not as an explicit external module). Due to such prompt design, CoT-TL generates output in a similar way, which makes it interpretable. In summary, we prompt the individual navigation steps and then prompt their relationship. Similarly, prompts are extended for higher-order relationships such as 3-ary, 4-ary, etc. \\
\textbf{LLM Input:} Natural language specification ($\mathcal{U}$), instruction regarding LTL translation, SRL of $\mathcal{U}$ as described in~\ref{sub:SRL}. \\
\textbf{Additional LLM Input}: AP identification from SRL, subgoal selection of $\mathcal{U}$ and progressively solving them into final LTL (as CoT). Fig.~\ref{fig:myprompt}, shows complete input provided to the LLM. \textit{Without CoT, at this stage, only the final LTL would have been provided, devoid of information as we provided above.} During testing, only the test specification $\mathcal{U}$ is appended to the few-shot CoT prompts examples without any additional information (such as chains-of-thought). \\ 
\textbf{LLM Output}: LTL output and reasoning chains involved in output generation (as learned from prompts).
\subsection{ Output Modal Checking \& Constraining}
\label{constrain}
LLMs suffer from hallucinations (incorrect or nonsensical formula)~\cite{gpt4} alongside uncertain output. To reduce it, CoT-TL employs the model checking library \textbf{Spot}~\cite{spot2}. In model checking, we verify that if LTL $\varphi$ can be satisfied by a finite state space model $M$ in time $O\left(\|M\| \times 2^{O(|\varphi|)}\right)$ or in space $O\left((\log \|M\|+\|\varphi\|)^2\right)$ complexity. For this checking, we checked whether generalized Büchi automata can be realized for the $\varphi$ (automata can be thought of as a state machine). In summary, $\varphi$ is a valid LTL generation iff its negation $\neg\varphi$ is not satisfiable. Spot checks this satisfiability~\cite{LTL_satisfy} and alerts us before accepting the $\varphi$, which can send the autonomous agent into an unsafe state. In case of unrealizable automata of $\varphi$, we re-prompt the LLM with the error message provided by the spot, incorporating the original prompt and the error message, to regenerate the LTL formula. Spot also helps CoT-TL to parse the LTL formula and generate a consistent desired output format (e.g., infix or prefix syntax). The above novelty can also be extended to a fine-tuning-based approach, and we believe it will improve the accuracy of other methods for LTL generation. 

The self-consistency check (\ref{sub:LLM}) module compares the $k$ (usually an odd number) different LLM-generated outputs, which the spot library accepts as valid LTL formulas. We use majority voting among the $k$ outputs to accept the final LTL specification, which can be used for planning. If majority voting is not possible, we take the most confident translation\footnote{Confidence score for each APs, logical and temporal operator is computed by average no. of appearance among $k$ outputs and based on this confidence score of each complete formula is calculated.}, but CoT-TL can be customized to generate an error message for such cases. Unlike~\cite{Wang2022SelfConsistencyIC}, who apply a consistency check to any output of the LLM, regardless of its automaton validity, we restrict our consistency check to only valid automaton formulas. Algorithm~\ref {algo} shows the complete interaction of all the modules for LTL generation.
\\
\textbf{Input to model checking}: LLM generated LTL. \\
\textbf{Model checking output}: either a) Correct automation: keep LLM-generated output directly. \textit{or} b) Incorrect automation: Reprompt the LLM with the error message.
\begin{algorithm}
\caption{CoT-TL:Natural Language Specification to LTL}
\label{algo}
\begin{algorithmic}[1]
\Require Dataset $\mathcal{D}$ consists of natural language instruction $\mathcal{U}$ and LTL translation $\varphi$. Prompt file $\mathcal{P}$ with few examples showing the derivation of $\varphi$ from $\mathcal{U}$, leveraging SRL of $\mathcal{U}$ in prompt and CoT linking $\mathcal{U}$ to $\varphi$. Model checking library \textit{spot} for checking Automaton $\mathcal{A}$ of $\varphi$.
\State Select large language model $LLM$
\ForAll{$\mathcal{U} \in \mathcal{D}$}
    \State $successRuns \gets 0$
    \State $results \gets \{\}$
    \While{$successRuns < 3$}
        \State $\mathcal{P} \gets \mathcal{P}$
        \State $input \gets \mathcal{U}$
        \State $\mathcal{P}.append(input)$
        \State $\varphi \gets LLM(\mathcal{P})$
        \State $AutomatonCheck \gets spot(\varphi)$
        \If{$AutomatonCheck \neq error$}
            \State $results.append(\varphi)$
            \State $successRuns \gets successRuns + 1$
        \Else
            \State $promptCopy \gets \mathcal{P}.copy()$
            \State $promptCopy.append(ErrorMessage)$
            \Comment{Rerun LLM for max 5 tries else failure for run}
            \State $\varphi \gets LLM(promptCopy)$
            \State $AutomatonCheck \gets spot(\varphi)$
                \State $results.append(\varphi)$
                \Comment{if rerun is error free}
                \State $successRuns \gets successRuns + 1$
        \EndIf
    \EndWhile
\State $\varphi \gets majorityVote(results)$
\EndFor
\end{algorithmic}
\end{algorithm}

\subsection{Demo: LTL-based Agent Planning}
The parsed and satisfiable LTL formula is used to synthesize robot trajectories. To demonstrate this, we simulated multistep drone navigation for a mobile quadcopter. The quadcopter has six states, each for 3D-pose and linear/angular velocity, and is modeled as a 12D double integrator. We compute the feasible trajectory using LTL as a constraint by solving mixed-integer convex programming~\cite{raman2014model} with the Gurobi~\cite{gurobi} solver. Inspired by Code-as-Policies~\cite{cop}, we leveraged LLM to encode the LTL formula as mixed-integer linear constraints on the continuous system variables. This transformation turned our planning problem into an optimization problem that Gurobi could solve, paving the way for converting LTL specifications into executable MATLAB code for robotics. See our demonstration video for a spoken language to drone planning example.

\section{Evaluations}
\label{eval}
\begin{table*}[ht!]
\begin{center}
\vspace{3pt}
\caption{ Translation accuracy (in \%). \colorbox{blue!20}{Baseline}, \colorbox{orange!20}{Ours}, \colorbox{green!20}{Ablation}. The \textit{Training Type} indicates the method of training or prompting, and \textit{full dataset} indicates that the model is tested on the complete dataset (without any holdout data). \vspace{-5pt}}
\begin{tabular}{l l l c c c c c c}
\hline
Model architecture & Training Type (Fine-tuning or Prompting) & Test data & Drone & CleanUp & Pick-Place\\
\hline
\rowcolor{blue!20}RNN~\cite{robotdataset}& Fine-tuning (synthetic and augmented data) & Full dataset & 22.41 & 52.54 & 32.39 \\
\rowcolor{blue!20}CopyNet~\cite{copynet} & Fine-tuning (synthetic and augmented data) & Full dataset & 36.41 & 53.40 & 40.36\\
\rowcolor{blue!20}BART-FT-Raw~\cite{dronetestfine} & Fine-tuning (synthetic, no augmented data) & Full dataset & 29.43  & 52.51 & 80.38 \\
\rowcolor{blue!20}BART-FT-Raw (or best BART-FT-Raw)~\cite{dronetestfine} & Fine-tuning (synthetic and augmented data) & Full dataset & 69.39  & 78.00 & 81.45 \\
\rowcolor{blue!20}Lang2LTL~\cite{lang2ltl} & 6-shot Prompting & Full dataset & 69.24  & 82.00 & 85.51 \\\hline
\rowcolor{orange!20}CoT-TL with GPT-4 (Ours) & CoT Prompting (6-shot)-No fine-tuning & Full dataset & \textbf{79.61 $\pm$ 1.3}  & \textbf{91.69$\pm$1.4} & \textbf{90.04$\pm$1.9} \\
\rowcolor{orange!20}CoT-TL with GPT-3 (Ours) & CoT Prompting (6-shot)-No fine-tuning & Full dataset & 61.25$\pm$1.2  & 73.10$\pm$1.3 & 77.05$\pm$2.2 \\ 
\rowcolor{orange!20}CoT-TL with Mistral-7b (ours) & CoT Prompting (6-shot)-No fine-tuning & Full dataset& 52.10$\pm$1.7  & 46.41$\pm$1.4 & 77.28$\pm$1.15 \\ 
\rowcolor{orange!20}CoT-TL with Starcoder (Ours) & CoT Prompting (6-shot)-No fine-tuning & Full dataset & 46.17$\pm$1.6  & 41.35$\pm$1.6 & 70.38$\pm$0.9 \\ \hline
\rowcolor{green!20}CoT-TL (Ours)-Classical CoT as~\cite{wei2023chainofthought} & CoT Prompting (6-shot)-prompt as per~\cite{wei2023chainofthought} & Full dataset &  58.40$\pm$2.3  & 69.63$\pm$1.8 & 71.67$\pm$2.1 \\ 
\rowcolor{green!20}CoT-TL (Ours)-No CoT & Prompting (6-shot)-No fine-tuning & Full dataset &  42.45$\pm$1.2  & 50.26$\pm$1.5 & 61.67$\pm$1.3 \\ 
\rowcolor{green!20}CoT-TL (Ours)-No SRL & CoT Prompting (6-shot)-No fine-tuning & Full dataset & 72.19$\pm$1.4  & 84.23$\pm$1.7 & 86.46$\pm$1.7 \\ 
\rowcolor{green!20}CoT-TL (Ours)-No model checking & CoT Prompting (6-shot)-No fine-tuning & Full dataset & 75.45$\pm$1.1  & 87.01$\pm$1.8 & 89.45$\pm$1.4 \\ 
\hline
\end{tabular}\vspace{-15pt}
\label{tab:evaluation}
\end{center}
\end{table*}
We evaluate the ability of CoT-TL to translate English specifications to LTL with much less data than fine-tuned models. We use three datasets for evaluation: \textit{drone planning}~\cite{dronedataset}, \textit{CleanUp World}~\cite{robotdataset}, and \textit{pick-and-place}~\cite{robotdataset}, which cover robotic and drone navigation tasks. We analyze the impact of CoT, SRL, and model checking on accuracy and discuss the main challenges for CoT-TL. We also assess CoT-TL’s generalization ability on a new unseen dataset from a different domain alongside the limitations.
\subsection{Evaluation Setup}
\label{sub:setup}
We evaluate four LLMs as the backbone for CoT-TL: GPT-4, GPT-3, Mistral-7B~\cite{mistral} and Starcoder~\cite{Li2023StarCoderMT}. We selected GPT-4 and GPT-3 for their state-of-the-art performance~\cite{gpt4}, Mistral-7B, and Starcoder for their open-source license. We use a VerbNet parser to extract SRL values, which is used for the prompt creation. LTL translation can be assumed to be a code-generation task~\cite{ltlcodetask}, and we selected Starcoder for its good performance among open-source models for code generation. We set the temperature parameter as $0.2$ ($0\le temprature\le 2$) to reduce the LLMs randomness but allow some diversity in the reasoning chains. We use 6-shot prompting with CoT to have at least one prompt for each unique LTL structure and consistency across the datasets. Unique LTL structure refers to a distinct formulation of LTL and logical operators, e.g., \texttt{F(A \& ! B)} and \texttt{!(A) U (B)} are two unique structures. Generally, higher \textit{shots} results in higher accuracy~\cite{wei2023chainofthought}, but the LLMs context window size limits the number of shot examples. We follow the same sentence structure as the original dataset and report the average translation accuracy and standard deviation $\sigma$ of CoT-TL over three runs. We use $\sigma$ to account for the LLM's non-determinism. Unless stated otherwise, we discuss CoT-TL with GPT-4 in the following subsections.

\subsection{Evaluation Datasets}
Table.~\ref{tab:dataset_stat} shows datasets and their properties. In the drone plan dataset, the drone needs to move or avoid location per natural language specifications. For example, instruction as ``avoid the red room until going to the second floor" and LTL as \texttt{$!(red\_room) U (second\_floor)$}. However, this dataset has some unclear semantics regarding specifications and vague AP grounding. For example ``go to orange room without going to another floor" where another floor is grounded as ``\texttt{$third\_floor$}". This is the most challenging dataset.
\label{sub:datáset}
\begin{table}[ht!]
\begin{center}
\vspace{-1pt}
\caption{ Evaluation Dataset Size and Property\vspace{-5pt}}
\begin{tabular}{l l l c c c c c }
\hline
\makecell{Dataset \\ Name} & \makecell{Total \\ Instructions} & \makecell{Unique LTL\\Formula Structure} & \makecell{LTL \\ Formulas}& \makecell{APs}\\
\hline
\makecell{Drone} & \makecell{6185} & \makecell{5} & \makecell{343} & \makecell{12} \\
\makecell{CleanUp} & \makecell{3382} & \makecell{6} & \makecell{39} & \makecell{6} \\
\makecell{Pick-and-place} & \makecell{744} & \makecell{1} & \makecell{5} & \makecell{5} \\
\hline
\end{tabular}\vspace{-8pt}
\label{tab:dataset_stat}
\end{center}
\end{table}

In the CleanUp dataset, the robot navigates through the rooms to move itself or objects based on instructions. Example instruction: ``move the robot through the yellow or blue small room and then to the green room" and its LTL: \texttt{F \& | \textbf{B} \textbf{Y} F \textbf{C}} with grounding dictionary: \{yellow room: \textbf{Y}, green room: \textbf{C}, blue room: \textbf{}\}. LTL formula can be written in different syntax, as \texttt{F \& | B Y F C} is prefix syntax whereas \texttt{F((B | Y \& F(C))} is infix syntax, but both indicate the same specification. The dataset has some unclear groundings for APs and a lot of noise. We use the modified version of this data as in~\cite{dronetestfine} for standardization.

The pick-and-place dataset instructions are semantically easier to follow, but the task is complex, and the LTL formula has a parse tree length of $5$, making it complex and less intuitive. Example instruction: ``Pick up all blocks except green ones and place them in the crate", LTL: \texttt{G \& U \textbf{S} ! \textbf{C} F \textbf{C}} and the grounding dictionary is: \{green: \textbf{C}, scan: \textbf{S}\}, also \texttt{G \& U S ! C F C} = \texttt{G((S U !C) \& FC)}.
\subsection{CoT-TL Experiment Parameter}
\label{appen:modelpara}
We used the same prompt for consistency for all types of LLM backbones. For non-GPT models, we used Hugging Face hosted models. For parameters, we used temperature as $0.2$, $max\_new\_tokens$ as 400 for the generation of LTL translation. For prompting, we indicate the end of a few-shot example with the keyword \textbf{FINISH}. Each natural language specification was translated $k=3$ times for self-consistency check to obtain the majority vote of three runs. $k$ can be any odd number, but we selected three based on API usage cost. If the LTL output fails the automaton check, we rerun the natural language input until we get the LTL with a valid automaton with a maximum of five tries. If, after five tries, we still have no valid LTL, we generate an error message for that run.
\subsection{Results and Discussions}
\label{sub:results}
Baseline work typically does not directly translate robot planning instructions into LTL according to the prompts, as done in CoT-TL. The most related work is Lang2LTL~\cite{lang2ltl}, which uses preprocessing to convert natural language instructions into referring expressions and lifted translation before the use of prompting for translation tasks. CoT-TL does not need such a process for inference except for formal logic-specific CoT prompt creation. Moreover, we use an off-the-shelf tool to reduce human intervention. We reimplemented Lang2LTL with the same example pairs and the same number of prompts as our model for standardization. Table~\ref{tab:evaluation} shows that CoT-TL surpasses state-of-the-art models in low data settings, achieving competitive accuracy on all datasets with only 6 data samples without intermediate translation. This makes our framework highly adaptable and suitable for new and custom robotics applications, enabling the automated generation of domain-specific planning specifications. We \textit{compare our method with BART-FT-Raw only in low data setting}, where it is trained on synthetic data and tested on the full dataset. Unlike the BART-FT-Raw main variant, our method does not require fine-tuning, which outperforms us in the range of 4 to 10\% when fine-tuned on 80\% of the data and tested on the 20\% remaining data.

CoT-TL outperforms the baseline models RNN~\cite{robotdataset}, CopyNet~\cite{copynet}, BART-FT-RAW~\cite{dronetestfine}, and Lang2LTL across all datasets in low data regime. CoT-TL also beats prompt-based Lang2LTL and fine-tuned BART-FT-RAW in low data settings, but its main advantage is its data efficiency. Our method does not need data augmentation, data preprocessing, and fine-tuning. CoT-TL achieves $79.61\%$ accuracy on the drone plan dataset, $10.22\%$ more than best BART-FT-RAW and $10.37\%$ more than Lang2LTL. On the cleanUp and pick-and-place datasets, CoT-TL attains $91.69\%$ and $90.04\%$ accuracy, surpassing Lang2LTL by $9.69\%$ and $4.53\%$, respectively and best BART-FT-RAW by $13.69\%$ and $8.59\%$, respectively. CoT-TL with GPT-3 performs well but slightly worse than the other baselines on all three datasets. However, CoT-TL with Starcoder and Mistral-7B, although better than RNN and CopyNet models, falls behind the baseline (Lang2LTL and BART-FT-RAW) and GPT-based models. A possible reason is the smaller model size and less complex knowledge captured by these open-source models during pre-training. This also shows that CoT-TL exploits large parameter size models better. These open-source models have fewer parameters than the GPT-based model. However, we observed an interesting result for the pick-and-place dataset, where Starcoder and Mistral-7B CoT-TL achieved very competitive accuracy compared to GPT -4-based CoT-TL. This could be due to the low diversity of LTL structures in the dataset (only one unique structure).

Ablation studies indicate that our CoT approach has the most impact on accuracy in CoT-TL. Without CoT, there is a severe drop in accuracy (in the range of $\approx30-40\%$) compared to our best CoT-TL model across the datasets, followed by the impact of SRL. The use of SRL information improved the accuracy of CoT-TL by $\approx3-8\%$ across datasets. SRL may have a modest impact because LLMs with CoT already capture the semantic structure of the sentences implicitly, and it is a soft constraint for LLMs. Automaton check via Spot has little impact; it only improves accuracy by $\approx1-4\%$ for GPT-4 based CoT-TL. We found that this is because most incorrect translations result from improper bracket nesting or incorrect AP due to vague specifications, which did not cause an error in the automaton generation. E.g., instead of \texttt{$F(C)\wedge G(!Y)$}, CoT-TL generates \texttt{$F\bigl(C \wedge G(!Y)\bigr)$}. Although model syntax checking does not increase translation accuracy much, ensuring that the final LTL formula satisfies the automaton properties for safe planning is important. Following CoT as introduced in~\cite{wei2023chainofthought}, we obtained $\approx18-22\%$ less accuracy than our CoT methodology as explained in Fig.~\ref{fig:myprompt}. These observations regarding the impact of CoT and SRL for better translation accuracy are consistent across all datasets.

\subsection{New Dataset Generalization and Limitations}
\label{result:holdout}
We evaluated CoT-TL’s generalization ability to the new dataset; we used a new dataset called \textit{OSM}~\cite{copynet}. This dataset comprises $556$ navigation instructions for $22$ cities with map landmarks and attributes. Examples include instructions like ``Stay away from Angel St and find bakery” and $\varphi$ as \texttt{$\bigl(G(! Angel\_St) \wedge F(bakery)\bigr)$}. OSM differs (partially) from drone plans and cleanUp datasets in terms of domain (closed vs. open navigation), instruction semantics or linguistics, LTL formula, and structures. It has $6$ unique LTL structures, $308$ unique formulas, and $12$ APs. We evaluated this dataset twice: first, using the same model configuration and prompts as in the drone plan dataset evaluation. We used the prompts and configuration from the cleanUp world dataset evaluation for the second experiment. In summary, the prompts and model configurations are copied from cleanUp and drone experiments, but the model is evaluated on the new OSM dataset. In this setup, CoT-TL achieved \textit{$64.5\pm0.7\%$} and \textit{$62.8\pm1.1\%$} accuracy on OSM using a drone and cleanUp setup, respectively. Most errors were for instructions with new LTL structures, as only $3$ of the $6$ LTL structures are similar to the drone/cleanUp dataset. Only $19.9\%$ of these new LTL structures were correctly translated by CoT-TL. This shows that the model can generalize to new semantics and LTL formulas but not so well to new LTL structures.

\begin{figure}[t]
    \centering
    \includegraphics[width=0.48\textwidth]{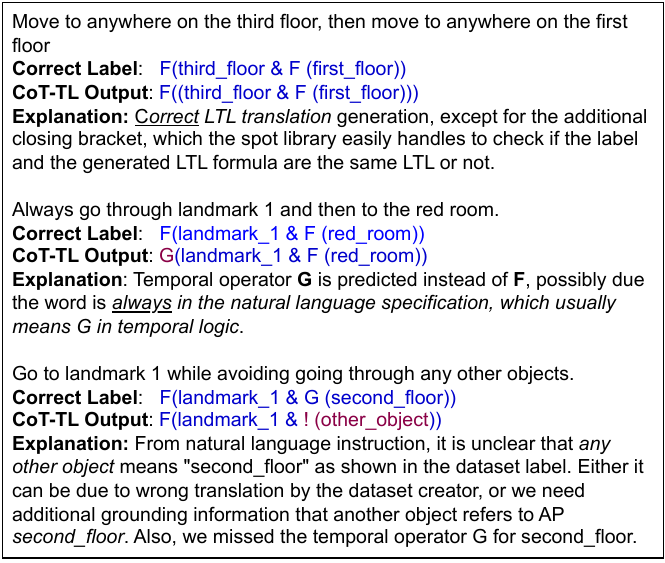}
    \vspace{-12pt}
    \caption{\textbf{Examples of CoT-TL Success and Failure} (explanations are manually written). The first example shows the successful case, and the rest demonstrates the failed cases.}
    \vspace{-16pt}
    \label{figs:failure}
\end{figure}

One possible limitation of our work is that comparing large model parameter LLMs to small ones may only be somewhat equitable. However, it is important to note that the observed improvements in LTL translation are not solely attributable to the larger model sizes. Instead, these enhancements result from several innovative approaches, including SRL, model checking, and the design of formal logic-specific CoT prompts that encourage coherent reasoning. Our ablation studies consistently revealed similar trends across all models, albeit with varying degrees of impact (we have omitted details on non-best-performing model ablations). Integrating these innovations even into the fine-tuned LLMs will significantly enhance their accuracy. Although we focused on easy-to-explain instructions as an example to explain our framework, Fig.~\ref{figs:failure} presents some successful and unsuccessful LTL translations. Also, we observed that CoT-TL is not as effective in LTL generation when the LTL structure is entirely different from that of prompts.
\section{Conclusion}
Our training-free framework, CoT-TL, translates natural language specifications into LTL for autonomous agents. This highly data-efficient method minimizes the need for a dataset for LLM fine-tuning. We show that with fewer, high-quality data samples, we can leverage LLMs' world knowledge to achieve high accuracy and generalization for LTL formalization tasks. CoT-TL offers interpretability and seamless adaptability to novel robotics tasks, ensuring swift deployment and integration into diverse application scenarios. On average, CoT-TL obtained $87\%$ accuracy across datasets with only six samples for each dataset. We attain this through CoT, SRL, and model-checking tool. Autonomous agents equipped with environment perception and localization can seamlessly leverage CoT-TL to plan trajectories based on natural language instructions. Closer integration of LLM and model checking can possibly address the current limitation of the new LTL structure generalization. The inherent uncertainty in LLMs necessitates exploration of its impact on planning, as the more complex the specification, the higher the likelihood of uncertain LTL structure.

\bibliographystyle{IEEEtran}
\bibliography{iros}
\addtolength{\textheight}{-12cm}   
\end{document}